\documentclass{article}
\usepackage{spconf,amsmath,epsfig,amsfonts}
\usepackage{multirow}
\usepackage{booktabs}
\usepackage[OT1]{fontenc}

\let\OLDthebibliography\thebibliography
\renewcommand\thebibliography[1]{
  \OLDthebibliography{#1}
  \setlength{\parskip}{0pt}
  \setlength{\itemsep}{0pt plus 0.3ex}
}

\begin{document}\sloppy

\def\x{{\mathbf x}}
\def\L{{\cal L}}

\title{Multi-object Tracking with A Hierarchical Single-branch Network}
%
\name{Fan Wang, En Zhu, Lei Luo, Siwei Wang, Jun Long}
\address{School of Computer, National University of Defense                       Technology, Changsha, China.\\
        wangfan10@nudt.edu.cn}

\maketitle

\begin{abstract}
Recent Multiple Object Tracking (MOT) methods have gradually attempted to integrate object detection and instance re-identification (Re-ID) into a united network to form a one-stage solution. 
Typically, these methods use two separated branches within a single network to accomplish detection and Re-ID respectively without studying the inter-relationship between them, which inevitably impedes  
the tracking performance. In this paper, we propose an online multi-object tracking framework based on a hierarchical single-branch network to solve this problem. Specifically, the proposed single-branch 
network utilizes an improved Hierarchical Online Instance Matching (iHOIM) loss to explicitly model the inter-relationship between object detection and Re-ID. 
Our novel iHOIM loss function unifies the objectives of the two subtasks and encourages better detection performance and feature learning even in extremely crowded scenes. Moreover, 
we propose to introduce the object positions, predicted by a motion model, as region proposals for subsequent object detection, where the intuition is that detection results and motion predictions 
can complement each other in different scenarios. Experimental results on MOT16 and MOT20 datasets show that we can achieve state-of-the-art tracking performance, and the ablation study verifies 
the effectiveness of each proposed component. 
\end{abstract}

\section{Introduction}
\label{sec:intro}

Multi-object tracking (MOT) is the basis of high-level scene understanding from video, which underpins significance application from video surveillance to autonomous driving. 
Recently, a few works have been proposed since the release of MOTChallenge \footnote{https://motchallenge.net/} which is the most commonly used benchmark for MOT. Generally, 
these works can be clustered into two categories: 1) \textit{two-stage} methods, namely \textit{tracking-by-detection} methods, that solve the problem of tracking multiple objects 
as two separate steps: object detection and data association; and 2) \textit{one-stage} methods that try to solve object detection, instance re-identification (Re-ID) and even data 
association in an end-to-end model. Commonly, two-stage methods \cite{bewley2016simple,wojke2017simple,tang2017multiple,xu2019spatial,long2018tracking} utilize a trained model to localize all instances 
in each video frame in the object detection step, and then, link detection results together to form object trajectories in the data association step. It means the MOT system requires at least 
two compute-intensive components: an object detector and a Re-ID model, which is intolerable for time-critical applications. One-stage methods 
\cite{bergmann2019tracking,zhou2020tracking,wang2019towards,zhang2020simple,zhang2020multiple,shuai2020multi} view MOT as a multi-task learning problem which avoid re-computation by sharing 
low-level features among different subtasks. In other words, how to effectively integrate different subtasks, such as object detection and Re-ID, into a single network is the key to ensure 
MOT performance.

\begin{figure}[t]
\centering
\includegraphics[width=0.48\textwidth]{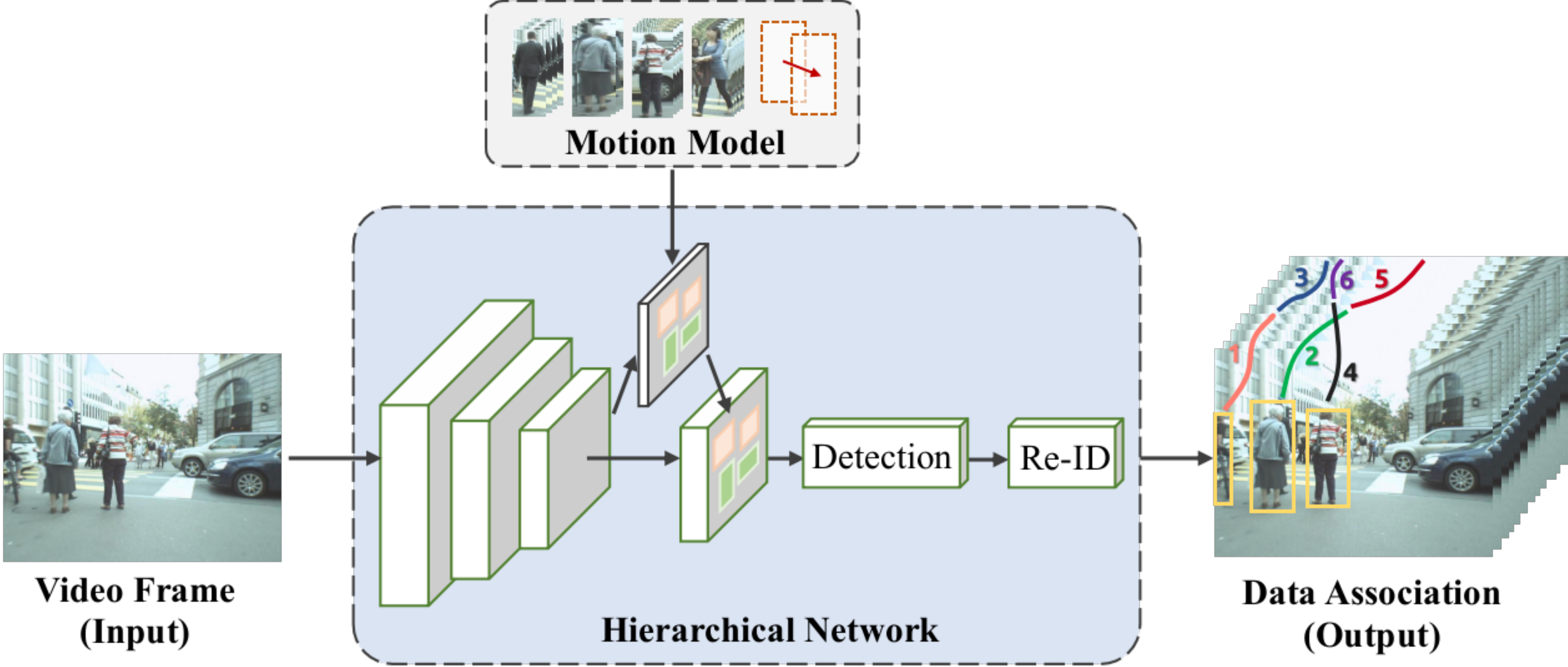}
\caption{Overview of our online multi-object tracking framework. The input video frames are fed into a hierarchical network to get object detection results and Re-ID features simultaneouly, 
in which motion information is integrated to complement the detection proposal process. Then, a data association operation is applied to get final object trajectories.}
\label{framework}
\end{figure}

As another important task in computer vision, person search aims to retrieve a person's image and corresponding position from an image dataset. Therefore, when the tracking object is a person, 
the tasks of MOT and person search share many similarities. The only difference is that MOT localizes and tracks objects on time-series video frames, while person search is based on a cluttered 
image dataset. In recent years, the research on person search has also undergone a progress from two-stage methods to one-stage methods. One-stage methods 
\cite{xiao2017joint,chen2018person,chen2020norm} greatly improve the speed of network inference while ensuring accuracy, but they suffer from contradictory 
objectives of detection and Re-ID. As disscussed in \cite{chen2018person}, simple concatenation of a detector and a linear embedding layer without harmonizing the two subtasks only leads to conflicting 
focusing points. In order to solve this problem, \cite{di2020hierarchical} proposes a hierarchical structure which explicitly models the relationship between pedestrain detection and Re-ID, and further 
exploits it as a prior to guide the one-stage model learning. The state-of-the-art performance in \cite{di2020hierarchical} shows that such a structure effectively integrates the commonness and uniqueness 
of pedestrain features and alleviates the contradictory objectives of detection and re-ID.

Inspired by the work above, we propose a hierarchical network combining motion information for multi-object tracking. As shown in Figure~\ref{framework}, each frame of a video is fed into a hierarchical 
network to get object detection results and Re-ID features simultaneouly. Then, data association utilizing spatial information and Re-ID features is applied to get final object trajectories. 
Specifically, the proposed single-branch deep network contains a hierarchical structure with two special layers, which are designed for detection and Re-ID respectively. The first layer captures the human 
\textit{commonness} and distinguish person from background, the second layer aims to classify persons' identities according to their \textit{uniqueness}. We improve the Hierarchical Online 
Instance Matching (HOIM) loss from \cite{di2020hierarchical} to explicitly formulate the inter-relationship between pedestrain detection and Re-ID. Moreover, we fuse the object motion predictions 
and region proposed network (RPN) outputs together as object region proposals. In this way, the hierarchical network not only uses the object appearance features, but also integrates the time-series 
information. 

In summary, our contribution is three-fold. First, a hierarchical single-branch network is proposed to explicitly model the inter-relationship between pedestrain detection and Re-ID.
Second, we introduce object time-series information (object motion model) into the hierarchical single-branch network to better localize objects. Third, we propose a one-shot framework for multi-object 
tracking and achieve state-of-the-art performance.

\section{related work}
\subsection{Joint Detection and Tracking}
A recent trend in multi-object tracking is to combine detection and tracking into a single framework, namely one-stage methods. Specifically, there are two ways: one is to combine detection 
and Re-ID into a single network to localize objects and extract appearance features simultaneouly; the other is to convert a object detector into a tracker directly.

\textbf{One-stage methods with Re-ID.} Wang et al. \cite{wang2019towards} formulate MOT as a multi-task learning problem with multiple objectives, such as anchor classification, 
bounding box regression and embedding learning, and reports the first near real-time MOT system. Zhang et al. \cite{zhang2020simple} study the essential reasons of the degraded results when 
attempting to accomplish detection and Re-ID in a single network, and presents a simple baseline to addresses this problem. Shuai et al. \cite{shuai2020multi} propose a detect-and-track framework, 
namely Siamese Track-RCNN, which consists of three functional branches: the detection branch, the Siamese-based track branch and the object re-identification branch. Peng et al. \cite{tai2020chained} 
propose a simple online model named Chained-Tracker, which naturally integrates object detection, object embedding and data association into an end-to-end solution. Although these methods try to use a 
single-shot deep network to accomplish detection and Re-ID, they do not essentially study the conflict between the two tasks. Instead, our proposed hierarchical structure explicitly formulates the 
relationship between detection and Re-ID, which is simple and efficient.

\textbf{One-stage methods without Re-ID.} Bergmann et al. \cite{bergmann2019tracking} directly propagate identities of region proposals using bounding box regression to realize data association. 
Zhou et al. \cite{zhou2020tracking} propose a simple online model named CenterTrack which only associates objects in adjacent frames, without reinitializing lost long-range tracks. 
Zhang et al. \cite{zhang2020multiple} design an end-to-end DNN tracking approach with two efficient trackers: FlowTracker and FuseTracker. The FlowTracker explores complex object-wise 
motion patterns and the FuseTracker refines and fuses objects from FlowTracker and detectors. These methods achieve a compromise between tracking speed and accuracy without using Re-ID features.

\subsection{Fusion of Detection and Motion Prediction Results}
How to effectively fuse results from motion model and detectors is also the key to improving the tracking accuracy. Zhang et al. \cite{zhang2018integrated} integrate the detection and tracking 
more tightly by conditioning the object detection in the current frame on tracklets computed in prior frames. In this way, the object detection results not only have high detection response, but also 
benefit a lot from existing tracklets. Feichtenhofer et al. \cite{feichtenhofer2017detect} link the frame level detections based on the across-frame tracklets to produce high accuracy detections at 
the video level. Chen et al. \cite{long2018tracking} present a novel scoring function based on a fully convolutional neural network to select a considerable amount of candidates from detection and motion 
model results. The major motivation for this is that detection and tracks can complement each other in different scenarios. Shuai et al. \cite{shuai2020multi} integrate a Siamese-based 
single object tracker into their proposed Track-RCNN which is robust to appearance changes and fast motion. Although these methods can fuse detections and tracks to a certain extent, they are usually 
complex and high computation. In our MOT framework, the motion model produces another kind of object region proposals which can be naturally merged into the proposed hierarchical network.

\begin{figure}[ht]
\centering
\includegraphics[width=0.48\textwidth]{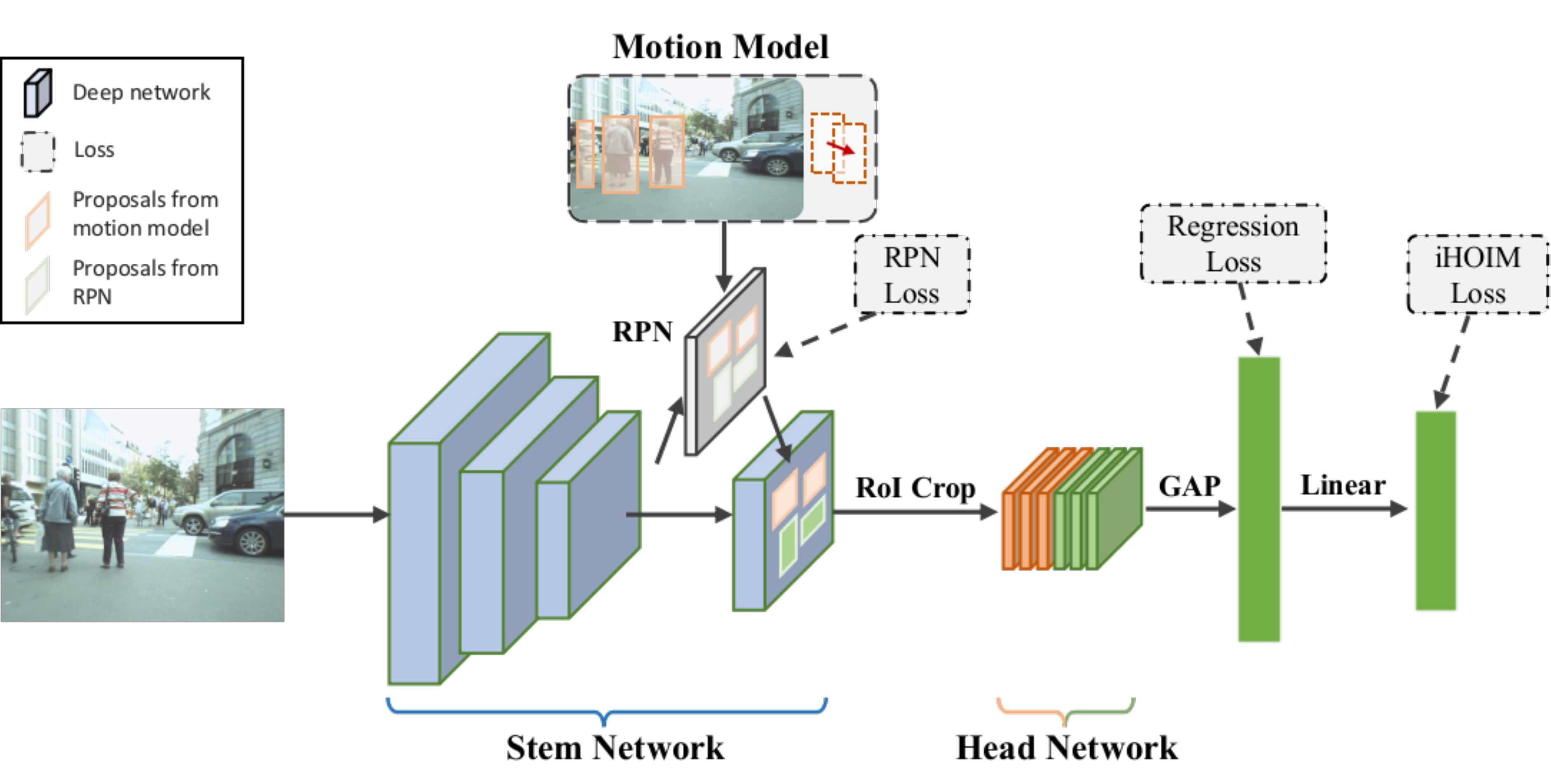}
\caption{Overview of our hierarchical single-branch network. Our network is based on Faster R-CNN \cite{ren2015faster} with a ResNet-50 \cite{he2016deep} backbone. In order to further improve the accuracy of object 
detection, we put the predictions of motion model as prior knowledge into the hierarchical network during inference, and fuse the PRN outputs and predictions together as object region proposals for 
subsequent box regression and feature extraction.}
\label{network}
\end{figure}

\section{Proposed Method}
In this work, we propose an online multi-object tracking framework with a hierarchical single-branch network which is shown in Figure~\ref{framework}. When a video frame is fed into the hierarchical single-branch network, we can 
get all object detection results and corresponding Re-ID features. Then, we obtain the object trajectories using a DeepSort \cite{wojke2017simple} framework. In this section, we first present 
an overview of the novel hierarchical single-branch network in Section \ref{h-network}. Then, we describe the improved Hierarchical Online Instance Matching (iHOIM) loss in Section \ref{hoim-loss} which explicitly formulates 
the relationship between detection and Re-ID.

\subsection{Hierarchical Single-branch Network} \label{h-network}
As shown in Figure~\ref{network}, the proposed hierarchical single-branch network is based on Faster R-CNN \cite{ren2015faster} with a ResNet-50 \cite{he2016deep} backbone which is composed of a stem network for 
sharing feature learning, a region proposal network (RPN) for generating object proposals, a motion model for object position prediction, and a head network (R-CNN) for box regression. At the end of 
the network, an extra $L_2$-normalized linear layer is added upon the top of the head network to extract object Re-ID feature.

During training, we remove the motion model from the hierarchical single-branch network. Following the configurations in \cite{di2020hierarchical}, we train the whole 
network jointly using Stochastic Gradient Descent (SGD) together with RPN loss (including proposal classification and regression loss), R-CNN box regression loss and the proposed iHOIM loss. 

During inference, we firstly feed the input video frame denoted as $\mathbf{I}_{t} \in \mathbb{R}^{3 \times w \times h}$ into the hierarchical single-branch network, a series of regions of interest 
$\mathbf{R}_{t} = \{r_t^1, r_t^2,..., r_t^l\}$ can be obtained at RPN layer. Simultaneously, a motion model is applied to predict object positions $\mathbf{M}_t = \{m_t^1,m_t^2,...,m_t^s\}$ 
in current frame based on the existing trajectories $\mathbf{T}_t$. Secondly, we fuse the boxes $\mathbf{R}_t$ and $\mathbf{M}_t$ as object region proposals $\mathbf{P}_t = \{r_t^1,...,r_t^l,m_t^1,...,m_t^s\}$ 
which will be fed into the head network. Finally, we can get object detection results $\mathbf{B}_{t} = \{b_t^1,b_t^2,...,b_t^n\}$ and Re-ID features $\mathbf{F}_{t} = \{f_t^1,f_t^2,...,f_t^n\}$ corresponding to the 
input video frame $\mathbf{I}_{t}$ at the last two layers of the network.

\subsection{Improved Hierarchical Online Instance Matching Loss} \label{hoim-loss}
At present, the one-stage networks \cite{wang2019towards,shuai2020multi,tai2020chained} commonly use two separated branches to accomplish detection and Re-ID based on the extracted sharing feature maps. 
These two branches methods do not study the competition of the two subtasks which inevitably impedes the tracking performance. In order to solve this problem, Chen et al. \cite{di2020hierarchical} propose 
the HOIM loss which is meant to integrate the hierarchical structure of person detection and Re-ID into the OIM \cite{xiao2017joint} loss explicitly. HOIM loss constructs three different queues to store 
labeled person, unlabeled person and background embeddings. However, it is not appicable in MOT scenario.

\begin{figure}[t]
\centering
\includegraphics[width=0.48\textwidth]{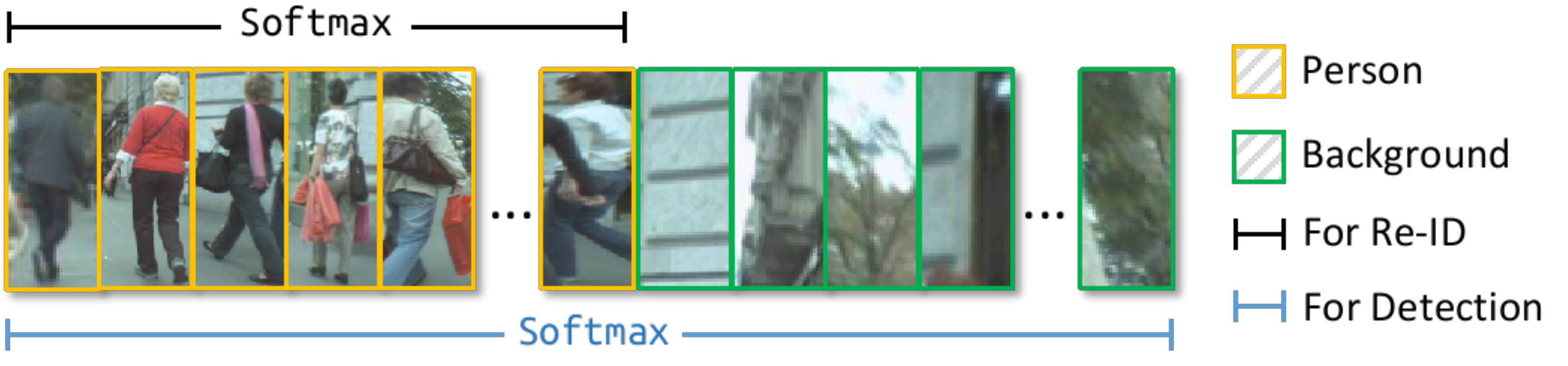}
\caption{The hierarchical structure that describes the inter-relationship between object detection and Re-ID. All the training losses are formulated on the law of total softmax probability 
        (black for detection loss and blue for Re-ID loss).}
\label{loss}
\end{figure}

In order to train more effectively on the MOT dataset, we propose an improved Hierarchical Online Instance Matching loss. Suppose there are $N$ different identities in the training data, iHOIM 
constructs a look-up table with size $N \times d$ to memorize the labeled person embeddings and a circular quene with size $B \times d$ to store a number of background embeddings. 
Together the look-up table and circular queue forms a projection matrix $\mathbf{W} \in \mathbb{R}^{(N+B)\times d}$. Given a proposal embedding $\mathbf{x}\in \mathbb{R}^d$, we can get the cosine distance 
between $\mathbf{x}$ and the stored embeddings by calculating a linear projection as follows: 
\begin{eqnarray}
\begin{array}{l}
~\mathbf{s}=\mathbf{W} \mathbf{x} \in \mathbb{R}^{N+B}, \\
\text { where } \mathbf{s}=\left[s_{1}, s_{2}, \ldots, s_{N}, s_{N+1}, \ldots, s_{N+B}\right],
\end{array}
\end{eqnarray}\label{equ:1}
then the probability of $\mathbf{x}$ belonging to an arbitrary person or background can be calculated by a softmax fuction:
\begin{eqnarray}
  p_i=\frac{e^{s_{i} / \tau}}{\sum_{j=1}^{N+B} e^{s_{j} / \tau}},
\end{eqnarray}\label{equ:2}
where $\tau$ is the temperature factor to control the softness of the probability distribution. Then the hierarchical structure that describes the  inter-relationship between object detection and Re-ID can 
be formulated on the law of total probability, which is shown in Figure~\ref{loss}.

For the first level of iHOIM loss for detection, we firstly consider the probability of an arbitrary embedding $\mathbf{x}$ that represents a person (denoted as $\Lambda$):
\begin{eqnarray}
  p(\Lambda)=\sum_{i=1}^{N} p_{i}.
\end{eqnarray}
Then, the probability of $\mathbf{x}$ represents background (denoted as $\Phi$) could be formulated in the same manner:
\begin{eqnarray}
  p(\Phi)=\sum_{i=N+1}^{N+B} p_{i}.
\end{eqnarray}
Combining these two probabilities, we formulate the object detection loss as a binary cross entropy loss:
\begin{eqnarray}
  \mathcal{L}_{\mathrm{det}}=-y \log (p(\Lambda))-(1-y) \log (p(\Phi)),
\end{eqnarray}
where $y$ is a binary label which equals 1 if $\mathbf{x}$ is a person, otherwise equals 0. 

For the second level, we follow \cite{xiao2017joint} to formulate the OIM loss for Re-ID. Given an embedding $\mathbf{x}$, the probability of $\mathbf{x}$ being a person and belonging to identity 
$k$ (denoted as \text{ id }$=k$) can be produced by a softmax function:
\begin{eqnarray}
  p(\text { id }=k, \Lambda)=\frac{e^{s_{k} / \tau}}{\sum_{j=1}^{N} e^{s_{j} / \tau}}.
\end{eqnarray}
Then, the objective of instance re-identification is to maximize the expexted log-likelihood:
\begin{eqnarray}
  \mathcal{L}_{\mathrm{OIM}}=\mathbb{E}_{\mathbf{x}}[\log p(\mathrm{id}=k, \Lambda)], \quad k=1,2, \ldots, N.
\end{eqnarray}

Finally, our proposed iHOIM loss is the linear combination of the two-level losses:
\begin{eqnarray}
  \mathcal{L}_{\mathrm{iHOIM}}=\mathcal{L}_{\mathrm{det}}+\lambda \mathcal{L}_{\mathrm{OIM}}, ~~\text{where~} \lambda=2p(\Lambda)^2,
\end{eqnarray}
where $\lambda$ is the loss weight for $\mathcal{L}_{\mathrm{OIM}}$. It dynamically weighs the importance of the two tasks based on detection confidence $p(\Lambda)$. The model focuses on identifying the 
detected person when the detection score is high, or focuses on detection task. By removing the circular quene for unlabeled person in HOIM \cite{di2020hierarchical} loss, our iHOIM loss is 
simpler and occupies less memory. It is not only able to identify different persons, but also classifies person from cluttered background. Thus, the embeddings are more robust and the detections 
are more accurate.

During training, the look-up table is update with a momentum of $\eta$:
\begin{eqnarray}
  \mathbf{w}_{k} \leftarrow \eta \mathbf{w}_{k}+(1-\eta) \mathbf{x}, ~~~~\text{if~} \mathbf{x} \text{~belongs to identity~} k,
\end{eqnarray}
and the circular quene replaces old embeddings with the new ones to preserve a fixed size.

\section{Experiments}
\subsection{Experiment Setup}
\textbf{Datasets and Metrics.} To evaluate the performance of our proposed tracking method, we conduct sufficient experiments on MOT16 \cite{milan2016mot16} and MOT20 \cite{dendorfer2020mot20} 
datasets, which are different in tracking scenes. Specifically, MOT16 dataset contains 7 training video sequences and 7 testing video sequences which are filmed in unconstrained 
environments. MOT20 dataset contains 4 training video sequences and 4 testing video sequences. All sequences in MOT20 dataset are filmed in very crowded scenes in which the density can reach values 
of 246 pedestrains per frame. We adopt multiple metrics used in the MOTChallenge benchmark to evaluate the proposed method, including multiple tracking accuracy (MOTA) 
\cite{bernardin2008evaluating}, identification F1 score (IDF1) \cite{ristani2016performance}, identity switches (IDSw), false positives (FP) and false negatives (FN).

\textbf{Implementation Details.} During training, we use Stochastic Gradient Descent (SGD) optimizer with the target learning rate of 0.003 which is gradually warmed-up 
at the first epoch and decayed by a factor of 0.1 at the 16 epoch, to train our hierarchical network on the training set of MOT16 and MOT20 respectively. The whole training process 
converges at epoch 22 with the batch size of 3 when we train it on a single GeForce TITAN Xp GPU. The momentum $\eta$ and softmax temperature $\tau$ of iHOIM are set to 0.5 and 1/30. 
Sizes of the embedding buffers, $i.e. N ~\text{and}~B$, are set individually according to the trajectories in different training datasets. For MOT16, they are 517, 500; for MOT20, $N$ is set to 2332, and $B$ is set to 2000 
to balance the probability distribution. Also, we employ the Selective Memory Refreshment (SMR) method from \cite{di2020hierarchical} to update the look up table for 
labeled person embeddings and circular quene for background embeddings.

During inference and tracking, we introduce the DeepSORT \cite{wojke2017simple} framework to tracking multiple objects based on the extracted detection results and corresponding embeddings. 
We choose the Kalman Filter as motion model to predict object positions based on the existing trajectories, which will be fed into the hierarchical network and work as region proposals for 
subsequent object detection. 

\begin{table*}[t]
\begin{center}
\caption{Results on the MOT Challenge test set benchmark. Up/down arrows indicate higher/lower is better.}
\vspace{8pt}
\label{table:mot16-20-test}
\scalebox{0.9}{
\begin{tabular}{clcccccc}
\toprule
{\bf Mode}&{\bf Method}&{\bf One-stage}&{\bf MOTA(\%){$\uparrow$}}
&{\bf IDF1(\%){$\uparrow$}}&{\bf IDSw{$\downarrow$}}&{\bf FP{$\downarrow$}}&{\bf FN{$\downarrow$}}\\
\midrule
\multicolumn{8}{c}{MOT16}\\
\midrule
\multirow{4}{*}{Batch}&GCRA \cite{ma2018trajectory}&$\times$&48.2&48.6&821&\bf{5104}&88586\\
&LMP \cite{tang2017multiple}&$\times$&48.8&51.3&481&6654&86245\\
&HCC \cite{ma2018customized}&$\times$&49.3&50.7&\bf{391}&5333&86795\\
&MPN \cite{braso2020learning}&$\times$&\bf{55.9}&\bf{59.9}&431&7086&\bf{72902}\\
\midrule
\multirow{5}{*}{Online}&RAR16 \cite{fang2018recurrent}&$\times$&45.9&48.8&\bf{648}&\bf{6871}&91173\\
&MOTDT \cite{long2018tracking}&$\times$&47.6&50.9&792&9253&85431\\
&STRN \cite{xu2019spatial}&$\times$&48.5&\bf{53.9}&747&9083&84178\\
&KCF \cite{chu2019online}&$\times$&48.8&47.2&906&5875&86567\\
&\bf{Ours} &$\checkmark$&\bf{50.4}&47.5&1826&18730&\bf{69800}\\
\midrule
\multicolumn{8}{c}{MOT20}\\
\midrule
\multirow{3}{*}{Online}&SORT \cite{bewley2016simple}&$\times$&42.7&45.1&4470&\bf{27521}&264694\\
&MLT \cite{zhang2020multiplex}&$\times$&48.9&\bf{54.6}&\bf{2187}&45660&216803\\
&\bf{Ours}&$\checkmark$&\bf{51.5}&44.5&4055&38223&\bf{208616}\\
\bottomrule
\end{tabular}}
\end{center}
\end{table*}

\subsection{Experimental Results and Analysis}
\textbf{Evaluation on Test Set.} Experimental results in Table~\ref{table:mot16-20-test} show that our proposed method achieves state-of-the-art performance compared with other 
advanced two-stage methods. Even in an extremely crowded scene like MOT20 dataset, our method still performs excellent. It shows that the proposed hierarchical structure effectively 
unifies the two tasks of object detection and Re-ID. We beat the previous best tracker by 1.6\%/2.6\% MOTA on MOT16/MOT20 respectively, which owes to the efficacy of our hierarchical 
network. Moreover, our one-stage MOT framework have much lower computational complexity and is about 5 times faster than the listed two-stage methods. The reason for the high IDSw 
and FP is that our model focuses on improving the quality of detection and Re-ID, while pays less attention to optimize tracking. It may be a future research direction of this work.

\begin{table}[htb]
\begin{center}
\caption{Ablation study results on MOT20 test set. \textbf{MM:} Motion Model, \textbf{$\Delta$:} Value Increment.}
\vspace{8pt}
\label{table:mot20-ablation}
\begin{tabular}{c|cc|c}
\toprule
{\bf Method}&{\bf MOTA(\%){$\uparrow$}}&{\bf IDF1(\%){$\uparrow$}}&{$\Delta$}\\
\midrule
OIM-MM&49.2&41.4&\\
\midrule
iHOIM-MM&51.2&43.0&(+2.0, +1.6)\\
iHOIM+MM&51.5&44.5&(+0.3, +1.5)\\
\bottomrule
\end{tabular}
\end{center}
\end{table}

\textbf{Ablation Studies.} In order to demonstrate the effectiveness of different components of our tracking framework, we ablate the two main components: iHOIM loss and motion model on the MOT20 test set. 
We first realize a OIM \cite{xiao2017joint} loss based tracking framework without using motion model. It shares the same network structure with our proposed model except that it separates detection and 
Re-ID supervisions into two independent losses, namely R-CNN classification loss and OIM loss. Then we add the proposed iHOIM loss into the hierarchical network without using motion model as well. As shown in 
Table~\ref{table:mot20-ablation}, the iHOIM loss helps to improve MOTA/IDF1 by 2.0\%/1.6\%. After adding motion model into the whole framework, we can get another 0.3\%/1.5\% increment on MOTA/IDF1. 
The exceptional performance substantially verifies the effectiveness of all the components.

\section{Conclusion}
In this paper, we propose an online multi-object tracking framework based on a hierarchical single-branch network. Concretely, 
we introduce an improved Hierarchical Online Instance Matching loss which explicitly models the inter-relationship between object detection and Re-ID. Moreover, a motion model is integrated 
into the proposed hierarchical single-branch network to complement the detection proposal process which improves tracking performance a lot. Compared with the two-stage methods on MOT16 and MOT20 datasets, 
our model achieves a new state-of-the-art performance even in crowded tracking scenes.

\bibliographystyle{IEEEbib}
\bibliography{icme2021template}

\end{document}